\documentclass{article}

    \usepackage[preprint, nonatbib]{neurips_2020}

\usepackage[utf8]{inputenc} 
\usepackage[T1]{fontenc}    
\usepackage{hyperref}       
\usepackage{url}            
\usepackage{booktabs}       
\usepackage{amsfonts}       
\usepackage{nicefrac}       
\usepackage{microtype}      
\usepackage{subcaption}
\usepackage{graphicx}
\usepackage{caption}
\usepackage{adjustbox}
\usepackage{multirow}
\usepackage{array}
\usepackage[ruled,vlined,noresetcount]{algorithm2e}
\usepackage{xcolor}

\title{Sample Efficient Training in Multi-Agent Adversarial Games with Limited Teammate Communication}

\author{%
  Hardik Meisheri \\ TCS Research \\ Mumbai, India \\ \textit{hardik.meisheri@tcs.com}
  \And 
  Harshad Khadilkar \\ TCS Research, IIT Bombay \\ Mumbai, India \\
  \textit{harshad.khadilkar@tcs.com}  
}

\begin{document}

\maketitle

\begin{abstract}
  
  We describe our solution approach for \textit{Pommerman TeamRadio}, a competition environment associated with NeurIPS 2019. The defining feature of our algorithm is achieving sample efficiency within a restrictive computational budget while beating the previous years learning agents. The proposed algorithm (i) uses imitation learning to seed the policy, (ii) explicitly defines the communication protocol between the two teammates, (iii) shapes the reward to provide a richer feedback signal to each agent during training and (iv) uses masking for catastrophic bad actions. We describe extensive tests against baselines, including those from the 2019 competition leaderboard, and also a specific investigation of the learned policy and the effect of each modification on performance. We show that the proposed approach is able to achieve competitive performance within half a million games of training, significantly faster than other studies in the literature.
\end{abstract}

\section{Introduction}

Game environments with cooperation (teammates) and adversaries (opponents) provide excellent testbeds for co-development of strategic and tactical behaviour in Reinforcement Learning (RL) settings. At the same time, the computational effort required for learning competitive policies using pure RL can be prohibitive, when the environment has a dynamic nature and poses a variety of challenges such as partial observability, credit assignment, and sparse delayed reward.

Pommerman environment based on Bomberman game (Hudson Soft, 1983) and first introduced as a competition in 2018~\cite{resnick2018pommerman}, provides an opportunity to handle all of these challenges simultaneously. In this paper, we propose an ensemble of techniques previously used in other simpler environments, which help us deal with a tightly constrained computational budget. We believe that this combination of ideas can be used in general to work with complex environments. This paper discusses our approach for solving \textit{TeamRadio} variant of Pommerman, which allows limited communication between teammates. Our modifications to the RL process collectively achieve competitive performance against a variety of baselines within $5 \times 10^5$ games.

Pommerman team competition was first held in 2018, where tree-based methods such as MCTS and rule-based agents dominated the rankings~\cite{osogami2019real, zhou2018hybrid}. The complexity of an environment in search space can be approximately attributed to the breadth and depth of the trees. For games like Chess and Go, these can be approximated to $b^d$, where b denotes breadth and d denotes the length of the game. Typically for chess, it is $35^{80}$, and for GO it is $250^{150}$~\cite{silver2016mastering}, whereas for Pommerman with 6 actions and 800 time steps, it can go up to $6^{800}$, when we consider a 1-\textit{vs}-1 scenario. The complexity is even higher in a 2-\textit{vs}-2 team game, since we deal with a joint action space. Solving this problem size in combination with the challenges described in Sec 2, with vanilla RL algorithms and a limited budget is not feasible. We observe that RL algorithms fail to learn anything significant without any reward shaping and action masking and these findings are consistent with earlier literature~\cite{meisheri2019accelerating, gao2019skynet}, though there appears to be no prior study specifically on Pommerman TeamRadio. The closest studies we found were for Pommerman 2-\textit{vs}-2 without communication. Briefly, our contributions are:\\
a) Providing a framework to learn competitive policies in a complex environment such as Pommerman TeamRadio, within a constrained computational budget\\
b) Extensive analysis and experimentation providing insights into the various aspects of learned policy based on empirical evidence.

\section{Description of environment: Pommerman TeamRadio}\label{sec:Pommerman}

\begin{figure}[b]
	\centering
	\begin{subfigure}{.5\textwidth}
		\centering
		\includegraphics[width=.94\linewidth]{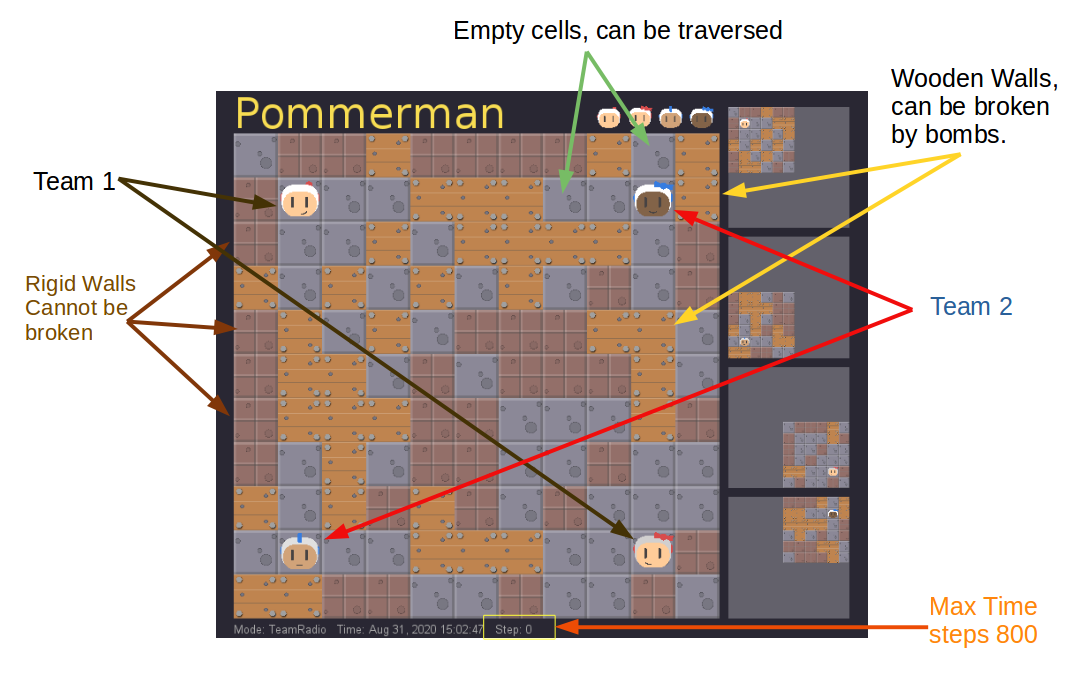}
		\caption{}
		\label{fig:sub1}
	\end{subfigure}%
	\begin{subfigure}{.5\textwidth}
		\centering
		\includegraphics[width=.97\linewidth]{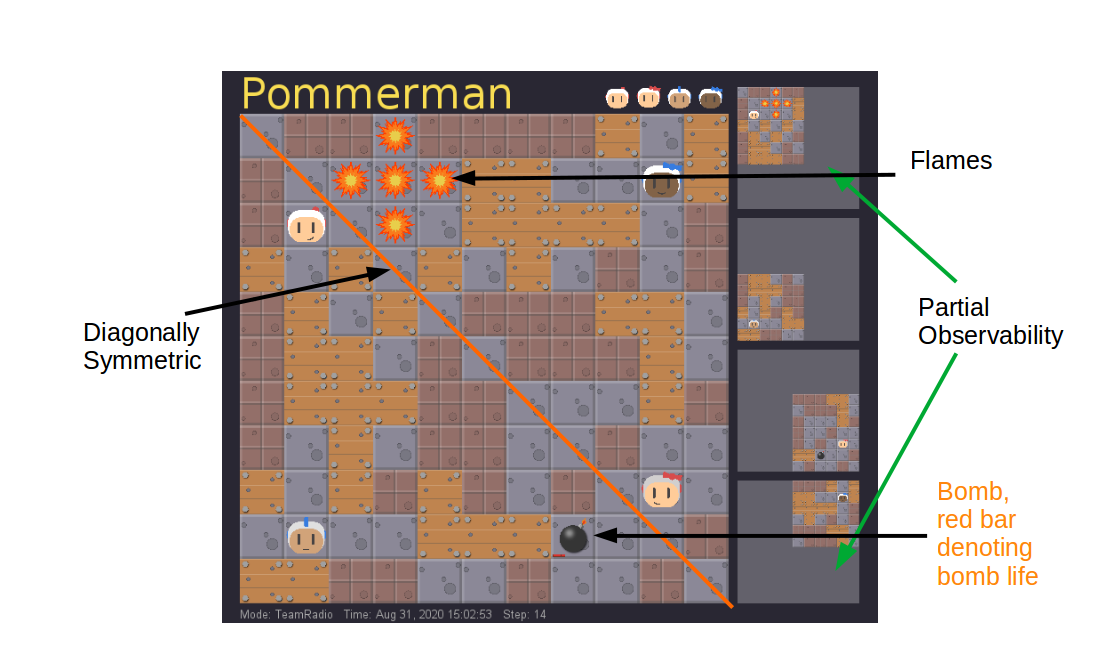}
		\caption{}
		\label{fig:sub2}
	\end{subfigure}
	\caption{Sample board of game}
	\label{fig:sample_board}
\end{figure}

Pommerman TeamRadio is a 2-\textit{vs}-2 board game as shown in Figure~\ref{fig:sample_board}. The 11$\times$11 board is randomly generated in each instance, including walls and passages (however, they are diagonally symmetric and there is always a route present from one agent to another). There are 4 agents with two teams, the teammates spawning in diagonally opposite corners. The board consist of passages which agents can walk through, there are two kinds of walls: wooden (breakable) and rigid (unbreakable). There are 6 actions: 4 cardinal actions (UP, DOWN, RIGHT, LEFT), BOMB placement and DO NOTHING. Each agent can see only 5 cell distance around itself, as seen in the inset panels next to the board. Agents start with a single bomb each with a blast radius of 3 cells, exploding 10 time ticks after placement. Wooden walls can be broken if they lie in the blast radius of any bomb and may reveal powerups (which give additional advantages if picked up, such as more bombs, larger blast radius, or the ability to kick a bomb away). To win a game, the team has to kill the enemy team within 800-time ticks. Game is deemed to be tied if it exceeds this time without a team winning (both enemies dead). Environment reward is +1 for winning, and -1 otherwise. 

Rules surrounding powerups and bombs lead to increased complexity in the game. Though each agent starts with the same abilities, these change through the game. Furthermore, the powerups appear randomly, so it is not possible to preplan their acquisition. The delayed bomb blasts and chaining of blasts (when multiple bombs cause a cascade of explosions) lead to an extremely dynamic environment. To summarise, the key challenges in this environment are: a) Variance in game length: Could be anything between 100 and 800 ticks, b) Sparse rewards at end of game lead to temporal credit assignment problem, c) Credit assignment among agents in a team cannot be accurately determined due to partial observability, bomb chaining and single reward for the team at the end of episode and d) Limited communication between agents creates a huge hindrance in planning actions as a team.

\textbf{Related Work:}\\
Top three approaches in the 2018 competition of Team variant were tree search based: HakozakiJunction, Eisenach, Dypm~\cite{pomm_book_chap, osogami2019real}. In the learning agent category, Navocado was the best learning performing agent followed by skynet~\cite{gao2019skynet}. Navocado used A2C agent and continual match based training (COMBAT) training framework. They show that during the course of training agent can learn different skills such as hiding, placing bombs etc. Skynet used PPO algorithm along with reward shaping for different actions such as picking of powerups and blasting wooden walls to solve the credit assignment. Pommerman has been used as an testbed for novel algorithms, where the goal is to solve/understand the a specific challenge instead of building a best performing agents~\cite{kartal2018using, kartal2019safer, perez2019analysis, malysheva2018deep}. In \cite{meisheri2019accelerating} they propose curriculum based learning along with imitation learning to solve the accelerate the training process in the complex environment as pommerman.We believe that \cite{gao2019skynet, meisheri2019accelerating} are closest to our work, however, we integrate and introduce many new ideas such as conservative reward shaping, playing against the pool of opponents and state-space representation. In addition, agents mentioned above are not dealing with TeamRadio variant of competition. We compare our approach with available agents or with statistics reported in earlier literature. We observe that our agent is able to beat both of the best performing agent from 2018 competition and earlier versions of PPO agents proposed. 

\section{Methodology}\label{sec:Methodology}

We describe our approach based on Proximal Policy Optimisation (PPO) with modifications to improve sample efficiency. First, we set the context in terms of observability and Markov properties.

\subsection{Preliminaries}

In a multiagent environment, agents interact simultaneously with the environment and with each other. This are formalized as a Markov game~\cite{littman1994markov} $\langle \mathcal{P}, \mathcal{S}, \mathcal{A}, \mathcal{T}, \mathcal{R} \rangle$, where $\mathcal{P} $ is set of $n$ players, $\mathcal{S}$ is set of states, $ \mathcal{A} $ is a collection of action sets, $\mathcal{A}_1, ... , \mathcal{A}_n$, $ \mathcal{T} $ is a state transition probability function, which provides probability of going from state $s$ to $s`$ given joint action of all players and $ \mathcal{R}: \mathcal{S} \times \mathcal{A} \rightarrow \mathbb{R}^n $ is a reward function. However, in Pommerman environment, each agent does not have complete information about the state including other players action, resulting in a Partially Observable MDP.

Decentralized MDPs (dec-MDPs), and decentralized POMDPs (dec-POMDPs) are some of the paradigms where it is no longer assumed that agents have perfect knowledge of the system state. However, solving dec-POMDPs with $n > 2 $ is nondeterministic exponential time complete \cite{bernstein2002complexity}. Using centralized control can help in solving the problem in polynomial time, but this is not possible with Pommerman. Therefore we model it as a MDP/POMDP setting $\langle \mathcal{S}, \mathcal{A}, \mathcal{T}, \mathcal{R}, \gamma \rangle$ where $\mathcal{S}$ represents the partially observable state along with belief state (details for belief state is provided in later section), $\mathcal{A}$ denotes the six actions, $\gamma $ is discount factor and $\mathcal{R}$ is reward for a single agent. 

\subsection{State Representation}

From the default state, we define an input size of $11 \times 11 \times 33$, where $11 \times 11$ is the size of the game board and 33 is the number of channels generated after encoding. Table~\ref{tab:state_space_representation} describes the encoding of the first 23 channels. We define an additional 10 channels for the `belief state' of the board, based on information known about all observed cells in the past 50 time steps. When one or more cells is under fog (not observable), we use the information from the belief state to fill them in, giving some global information (albeit noisy) about the entire board. We do this in non-parametric way to avoid creating additional learnable parameters. Importantly, we do not include agent positions in the belief state, given their highly dynamic nature. Some information about the agents is included in the communication messages, explained below.

\begin{table}[b]
	
	\caption{Current state representation (each channel of size $11\times 11$).}
	
	\begin{tabular}{m{3cm}m{10cm}}
		\hline
		Board representation & binary encoded channels for each of passages, rigid walls, wooden walls, bombs, flames, fog, extra bomb powerups, increase range powerups, kick powerups present on the board (10 channels in total) \\
		\hline
		Bomb meta information encoding & One channel each for encoding the bomb life remaining of the bomb and blast strength (2 channels)\\
		\hline
		Position encoding & One channel each for all four agents' positions (4 channels) \\
		\hline
		Alive encoding & Binary alive/dead state of the three agents apart from self (3 channels)\\
		\hline
		Powerups & One channel each for ammo, blast strength, kick capability (3 channels)\\
		\hline
		Time encoding & 11$\times$11 matrix filled with the value of the current time tick (1 channel)\\
		\hline
	\end{tabular}
	
	\label{tab:state_space_representation}
\end{table}

\subsection{Communication Protocol} \label{subsec:comm}

In TeamRadio variant of Pommerman, limited communication between the agents is allowed. At each timestep, an agent can send a maximum of two integers between 1 and 8 to its teammate, resulting in 64 unique symbols. In order to avoid the complexity of learning the communication protocol, we define rules to compute the symbol to be sent. The key information we transfer is positioning of self and enemies, since this is the most dynamic part of the game. We use only coarse encoding, since sending the exact coordinates is not feasible with 64 symbols. For both the enemies, we use 4 unique symbols to indicate the presence of that enemy in one of the quadrants of the board (if visible), and one symbol to indicate if you have no information regarding that enemy. This results in 5 symbols for each enemy position. The agent's own position is encoded with 2 symbols, indicating top- or bottom-half of the board. Thus we need $5\times5\times2$ unique symbols for communicating rough positions, utilising 50 out of the 64 combinations that can be sent. In the present version, the remaining 14 symbols are left unused. The communicated information is used to build the position encoding channels in Table \ref{tab:state_space_representation}, whenever the teammate or an enemy is not visible to the agent itself.

\subsection{Reward Shaping} \label{subsec:rewards}

As mentioned earlier, credit assignment is a particular challenge in Pommerman because of the long episode length, delayed and chained bomb explosions, and the presence of a teammate. In complex environments, reward shaping is typically not encouraged as it introduces unintended changes in the value function and can lead to suboptimal policies~\cite{ng1999policy, openAI_blog}. To minimise unintended consequences, we modify only the terminal reward. If one of the enemies dies before an agent's trajectory ends (whether through its death, or through the 800-step timeout), the agent gets a reward of $+0.5$. If both enemies are alive when an agent dies, it gets $-1.0$. If both enemies are dead, it gets $+1.0$. As the state representation consists of alive broadcasting channel, it can specifically relate to this reward structure and can create a meaningful representation. The rules are applied for each teammate independently. We refrain from assigning credit based on specific enemy kills by an agent, as it is very difficult to ascertain which bomb(s) killed which agent(s) due to partial observability and bomb chaining. 

\subsection{Action Filter}

Placing of bombs is highly correlated with losing and suicides~\cite{resnick2018pommerman, kartal2019safer} and hence the vanilla model-free RL leads to lazy-agent or camping behaviour, where an agent learns the conservative strategy of sitting still and not placing any bombs at all. In order to avoid inadvertent suicides during training, we use an action filter \cite{gao2019skynet} which does one-step look ahead and determines positions which are clearly leading to suicides (because of imminent bomb explosions). Besides it sometimes prevents the agent from placing a bomb when in blast radius of nearby bombs. We only mask risky actions at each time step based on action filter, but the actual action out of the remaining ones is chosen by the exploration/exploitation policy.

\subsection{Opponents}

Inspired from Alphastar \cite{alphastarblog}, we create a pool of opponents with varying degree of aggressiveness and skillset to help learn more generalized and robust strategies and avoid overfitting against single strategy. The agent pool is shown in Table~\ref{tab:opponents}, and against each pool, it shows \% of games played in a batch. We give the highest percentage to the static opponent as it forces the agent to venture out of its initial position and kill an agent. This helps in developing engaging behaviour rather than \textit{camping} which is exhibited by Cautious PPO agent~\cite{meisheri2019accelerating}, when learned using vanilla PPO. The primary reason for such behaviour is bomb placement is highly correlated with agent suicides~\cite{kartal2019safer, shahmulti}. Least weightage is given to heuristic agents (smart simple and smart simple nobomb) as these agents' logic can be exploited to diagonally block and kill the heuristic agent with their own bombs\footnote{This is a bug in Simple Agent. Experiments are available in supplementary material.}.   

\begin{table}[h]
	
	\caption{List of opponents and frequency of appearance in training.}
	\begin{tabular}{m{3.5cm}m{1.5cm}m{8cm}}
		\hline
		Agents              &\% in batch & Remarks                                                                                                                           \\ \hline
		Simple              & - & Heuristic agent provided by competition organizers, uses Dijkstra's algorithm for paths fixed rules for attack  \\ \hline
		Static              & 50 & Only action possible is stop                                                                                              \\ \hline
		Smart Simple         & 5 & Simple Agent supplemented by our action filter                                                                                            \\ \hline
		Smart Simple Nobomb & 5 & Simple Agent with action filter but bomb action is masked                                                               \\ \hline
		Neoteric          & 20 & Heuristic agent based on hierarchical policy. This agent was ranked 4th overall in 2019 TeamRadio competition\footnote{Accessed after contacting authors, Sumedh Gupte and Sanjay Bhat}                \\ \hline
		Cautious PPO Agent  & 10 & Cautious agent which was learnt using PPO without any reward shaping, described by authors in  ~\cite{meisheri2019accelerating}                         \\ \hline
		PPO Agent     & 10 & Agent trained with reward shaping and curriculum for the 2018 version, described in ~\cite{meisheri2019accelerating}                                                               \\ \hline
	\end{tabular}%
	\label{tab:opponents}
\end{table}

\subsection{Training scheme with imitation learning and RL} 

To mitigate the issue of initial slow learning due to environment complexity, we use imitation learning on a noisy expert to pre-train the policy weights for PPO~\cite{schulman2017proximal}. We use the model architecture as described in~\cite{meisheri2019accelerating}, with $3$ CNN layers for policy and value networks. We collected 48K samples by playing 12K games with all four players as smart simple (Table \ref{tab:opponents}). The samples collected were then augmented with communication messages based on the actual states, as described in Sec \ref{subsec:comm}. We then performed the imitation learning exercise. Results of training loss and validation loss are given in supplementary material. The imitation learnt policy was then used for seeding the RL policy.

During RL training, action filter is incorporated into policy to avoid the sure-shot catastrophic scenario (note that the imitation samples also are filtered, since we collect them from the smart simple agent). We mask out the actions which are bad by creating a binary vector indicating allowed actions for each state. The ratio between the old policy and new policy in PPO is defined as \[ r_t(\theta) = \frac{\pi_t(a_t/s_t)}{\pi_{t_{old}}(a_t/s_t)}\]
The one-hot encoding of action filter is multiplied with both $\pi_t$ and $\pi_{t_{old}}$. This ensures low variance between the two policies. We train RL with 128 games worth of samples as batch size, which further helps in stabilizing gradients from all the opponents.

\section{Results and Discussion}\label{sec:Results}

\begin{figure}[h]
\centering
\includegraphics[width=\linewidth]{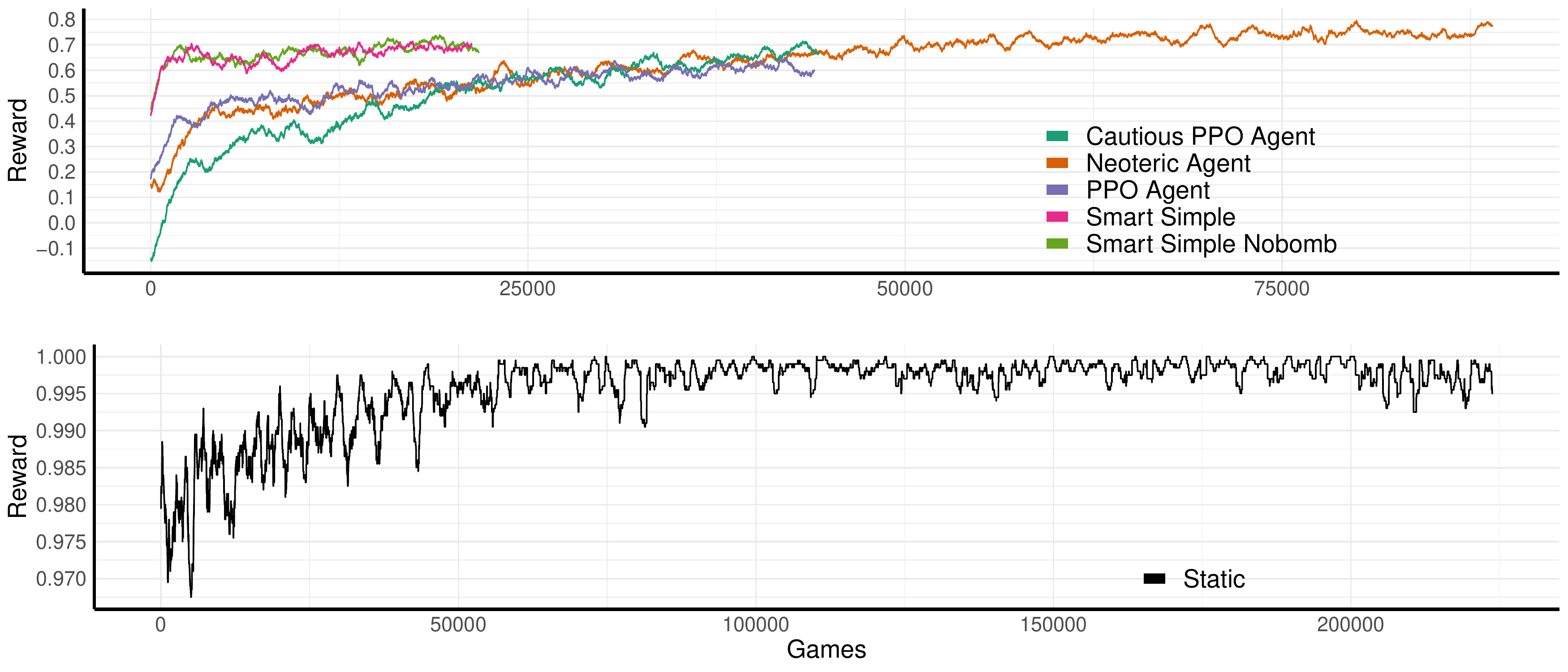}
\caption{Rewards during RL training (static agent is separate due to large difference in x-axis).}
\label{fig:training_results_wrt_opp}
\end{figure}

Figure~\ref{fig:training_results_wrt_opp} shows the improvement in shaped rewards (Sec \ref{subsec:rewards} during RL training, against each type of opponent described in Table \ref{tab:opponents}. The initial policy for RL is already trained using imitation learning, as described before. The reward is shown to increase and stabilise against each opponent. Table~\ref{tab:training_results} shows the win, loss and tie percentage improvement against all opponents, averaged over 1000 games each. As of this writing, we only had access to one agent (Neoteric) from the TeamRadio version for a direct comparison. Table \ref{tab:eval_simple_agent} and \ref{tab:eval_2018_best_Lragents} compares the performance of our agent against Simple Agent, along with top-ranked agents from previous years (versions without communication).

\begin{table}[h]
	\centering
	\caption{Improvement in results from imitation-learnt policy to that after RL. (SS = Smart Simple, SS n/b = Smart Simple with no bomb placement, and Cau. = Cautious)}
	\begin{tabular}{lrrr||rrr|||lrrr||rrr}
		\toprule
		&\multicolumn{3}{c||}{Imitation} & \multicolumn{3}{c|||}{Imitation + RL} & &\multicolumn{3}{c||}{Imitation} & \multicolumn{3}{c}{Imitation + RL}\\
		Opponent    &   W &   L &    T &   W &   L &    T & Opponent    &   W &   L &    T &   W &   L &    T       \\
		\midrule
		Static              &  21 &  2 &  77 &  96 &  1  &  3 & Neoteric            &  23 &  15 &  62 &  65 &  17 &  18 \\
		SS n/b &  53 &  12 &  35 &  73 &  16 &  11 & Cau. PPO  &  8  &  10  &  82 &  61 &  26 &  13 \\
		SS        &  53 &  11 &  36 &  73 &  15 &  12 & PPO           &  26 &  14 &  60 &  56 &  23 &  21 \\
		\bottomrule
	\end{tabular}
	\label{tab:training_results}
\end{table}

\begin{table}[h]
	\parbox{.45\linewidth}{
		\centering
		\caption{Results against Simple Agent compared to baselines (in \%)}
		\begin{tabular}{lrrr}
			\toprule
			Result &    Won &   Lost &    Tie \\
			\midrule
			PPO Agent~\cite{meisheri2019accelerating} &  77.8 &  8.7 &  13.5 \\
			MCTS~\cite{perez2019analysis} & 70 & 2 & 28 \\
			MAGnet~\cite{malysheva2018deep} & 71.3 & - & - \\
			Our Agent & \textbf{80.3} & 11.1 & 8.6 \\
			\bottomrule
		\end{tabular}
		\label{tab:eval_simple_agent}
	}
	\hfill
	\parbox{.45\linewidth}{
		\centering
			\caption{Against 2018 Best learning Agents (in \%)}
			\begin{tabular}{lrrr}
				\toprule
				Result &    Won &   Lost &    Tie \\
				\midrule
				Navocado~\cite{peng2018continual} &  47 &  29 &  24 \\
				Skynet~\cite{gao2019skynet} & 71 & 15 & 14 \\
				\bottomrule
			\end{tabular}
			\label{tab:eval_2018_best_Lragents}
	}
\end{table}

\subsection{Effect of reward shaping and action filter on rewards and map exploration}

\begin{figure}[h]
	\centering
	\begin{subfigure}{.5\textwidth}
		\centering
        \includegraphics[width=0.9\linewidth]{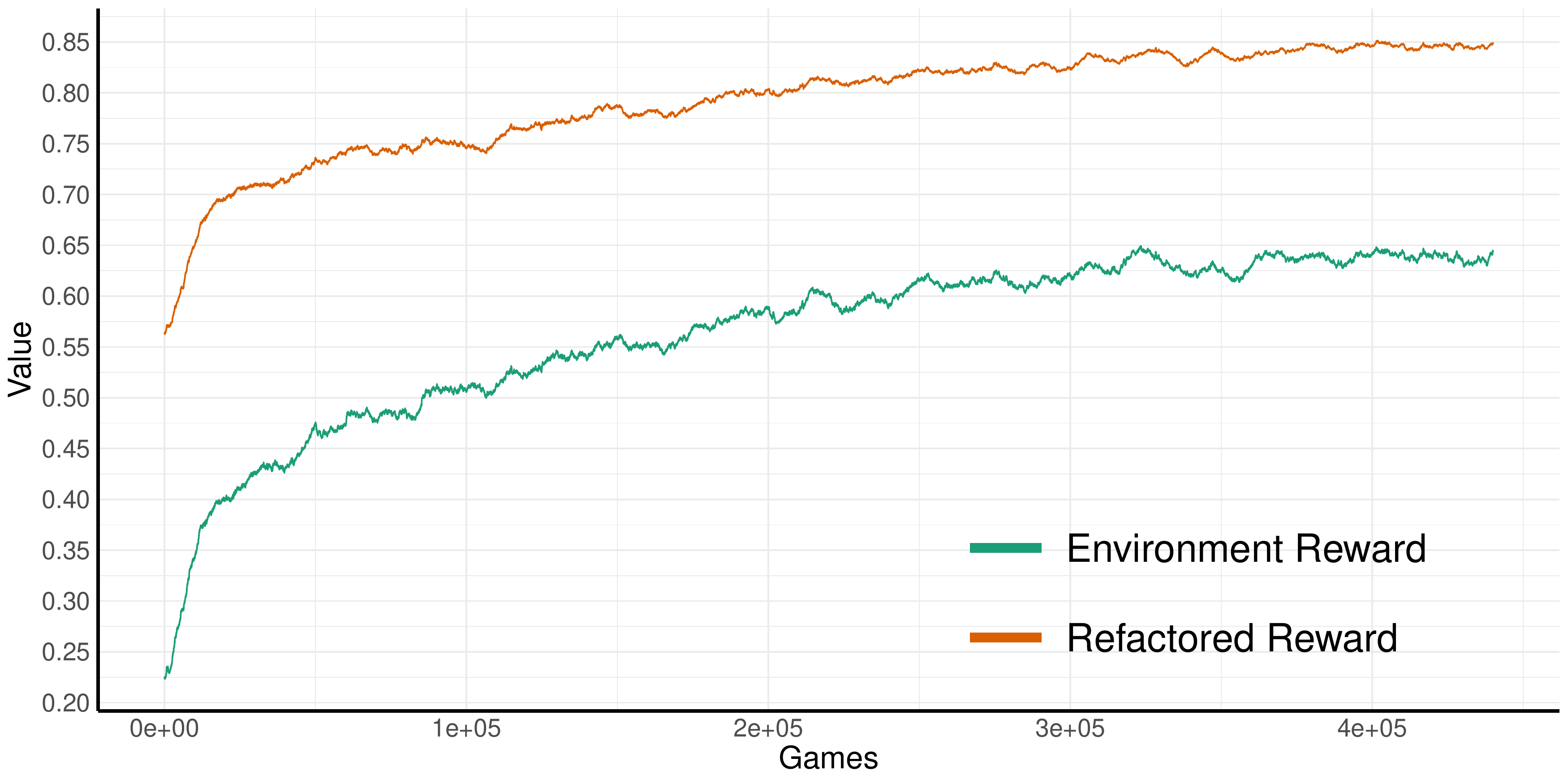}
        \caption{}
		\label{fig:training_results_overall}
	\end{subfigure}%
	\begin{subfigure}{.5\textwidth}
		\centering
        \includegraphics[width=0.9\linewidth]{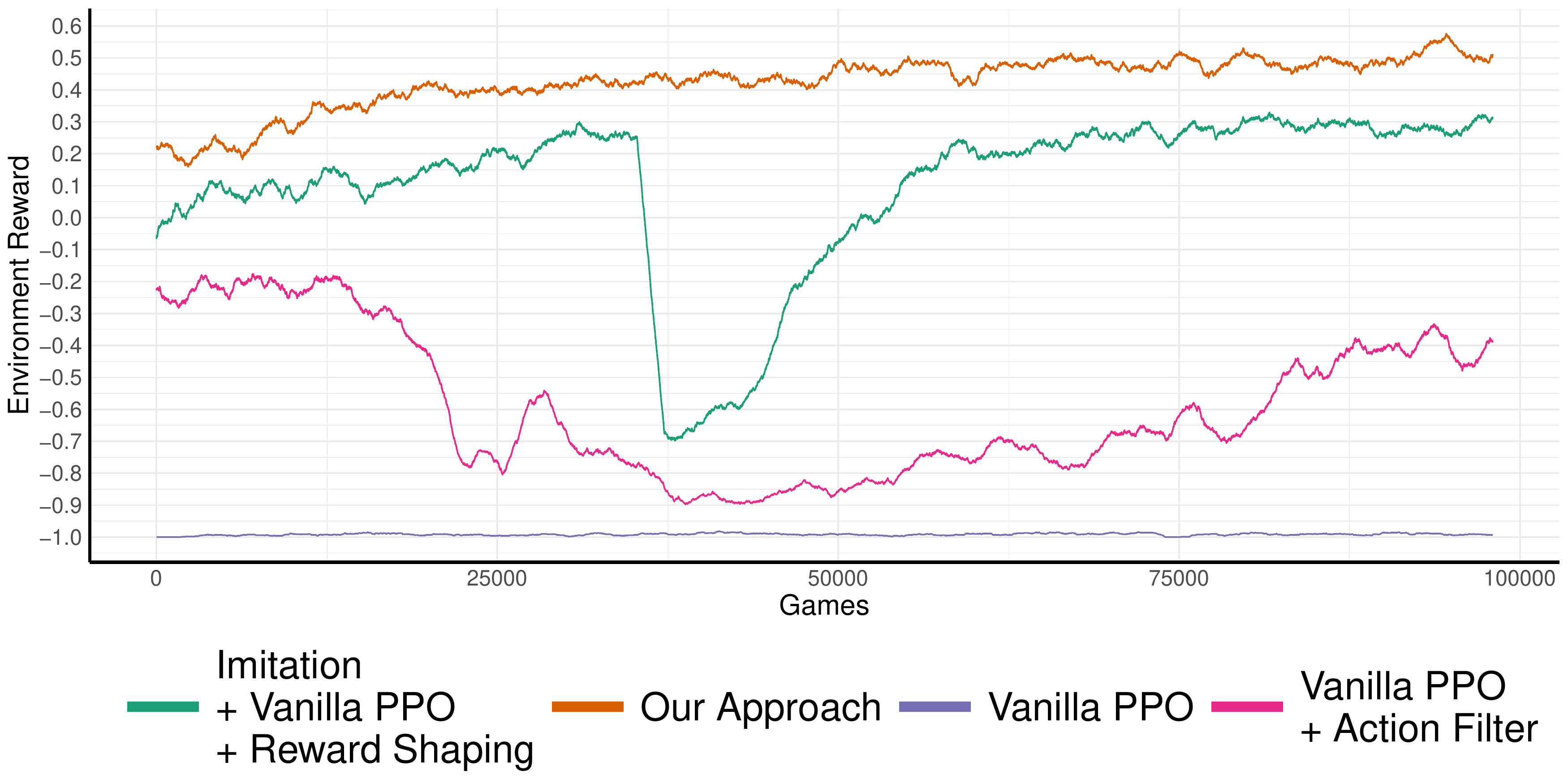}
        \caption{}
		\label{fig:different_training_mechs}
	\end{subfigure}
	\caption{a) Training results with 6 opponents, plotted with moving average of 10k, b) Effect of adding action filter and reward shaping. Y-axis represents the raw environment reward, and the graph is generated with moving average of 1k.}
	\label{fig:training_results_overall_1}
\end{figure}

Figure~\ref{fig:training_results_overall} plots the raw environment reward and our reshaped reward over the course of training. We see that there is a high correlation between the increase/decrease of reward in the two plots. To understand the effect of action filter and imitation on learning we ran training epochs for $1 \times 10^5$ games, Figure~\ref{fig:different_training_mechs} shows vanilla PPO applied to pommerman does not show any improvement in the expected rewards. Removing action filter from our approach leads to high variance in expected rewards as placing of bombs also leads to agent dying itself if there are no escape squares to be found. Having action filter with high value of gamma ensures that the policy is optimizing over long term goals whereas short term safety or immediate safety is taken care by the action filter. Without imitation it takes a long time to learn basic skills and exploration such as breaking of wooden walls, picking of powerups etc, the Vanilla PPO + action filter curve shows the effect of no reward shaping and imitation being present. The action filter is able to increase the chances of survival for longer duration however it still needs lot of time to learn.

\begin{figure}[h]
	\centering
	\includegraphics[width=0.75\linewidth]{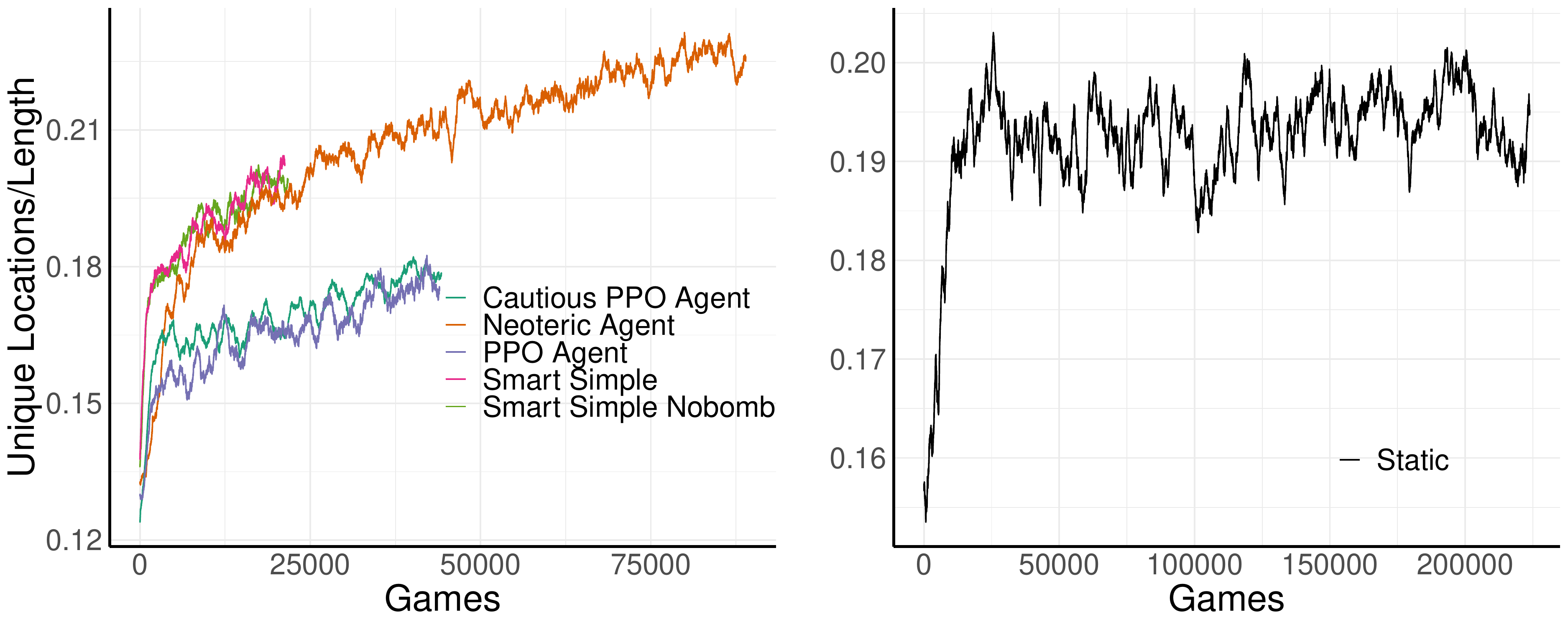}
	\caption{Number of unique cells visited per time step, over the course of training.}
	\label{fig:training_unique_cells_per_length}
\end{figure}

Figure~\ref{fig:training_unique_cells_per_length} shows the number of unique cells/tiles visited normalized by the time agent was alive in that game. This metric is proxy for understanding how the agent is exploring the space spatially in a given amount of time.


\subsection{Ablation studies of various factors on evaluation performance}

\begin{figure}[h]
	\centering
	\begin{subfigure}{.5\textwidth}
		\centering
		\includegraphics[width=0.98\linewidth]{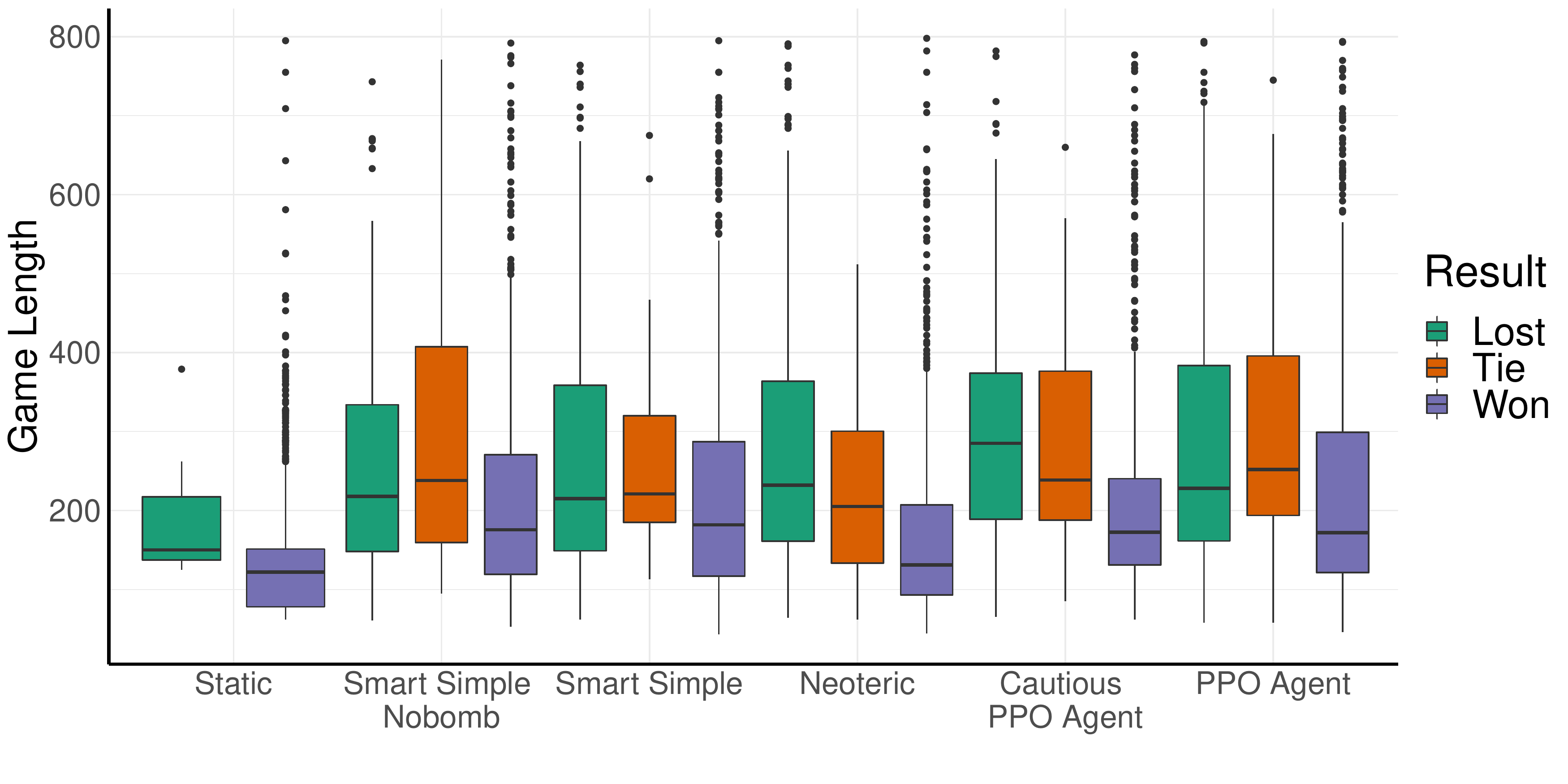}
		\caption{}
		\label{fig:game_length_distribution_across_opponents}
	\end{subfigure}%
	\begin{subfigure}{.5\textwidth}
		\centering
			\includegraphics[width=0.98\linewidth]{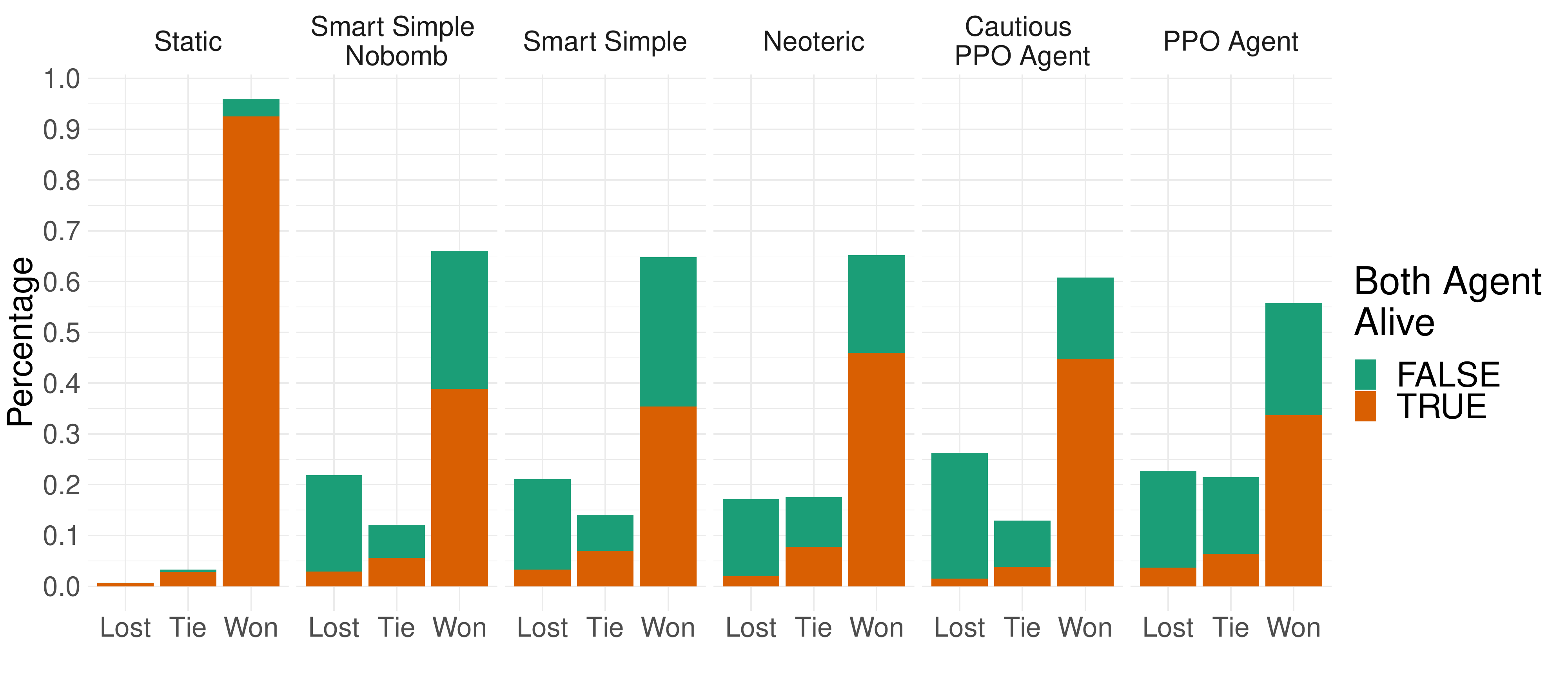}
		\caption{}
		\label{fig:result_distribution_teammate_death}
	\end{subfigure}
	\caption{a) Box plot of game length across different opponents (Games terminating due to timeout are not considered in this), b) Percentage distribution of results when one of the team mate dies across all the opponent}
	\label{fig:game_length}
\end{figure}

Figure~\ref{fig:game_length_distribution_across_opponents} shows a box plot of game length across different opponents, colored according to the end game result. We can observe that games that are won are typically shorter than others, across all opponents. We have removed the games which are being tied due to timeout that is 800 time ticks, as it would provide much more clear picture of understanding ties in which all the agents die simultaneously. Games against the static agent are shortest, since it is the easiest opponent.

A key component of team play is the effect of having an active teammate in the game. Figure~\ref{fig:result_distribution_teammate_death} shows how difference in win percentage when playing against all the six opponents, with and without the teammate. As expected, most of the losses are when the team mate has died, whereas a majority of games are won when the teammate is alive. Games against the static agent are predominantly with a teammate, since the only way the teammate can die is if the action filter does not mask an unsafe action.

\begin{figure}[h]
	\centering
	\includegraphics[width=0.7\linewidth]{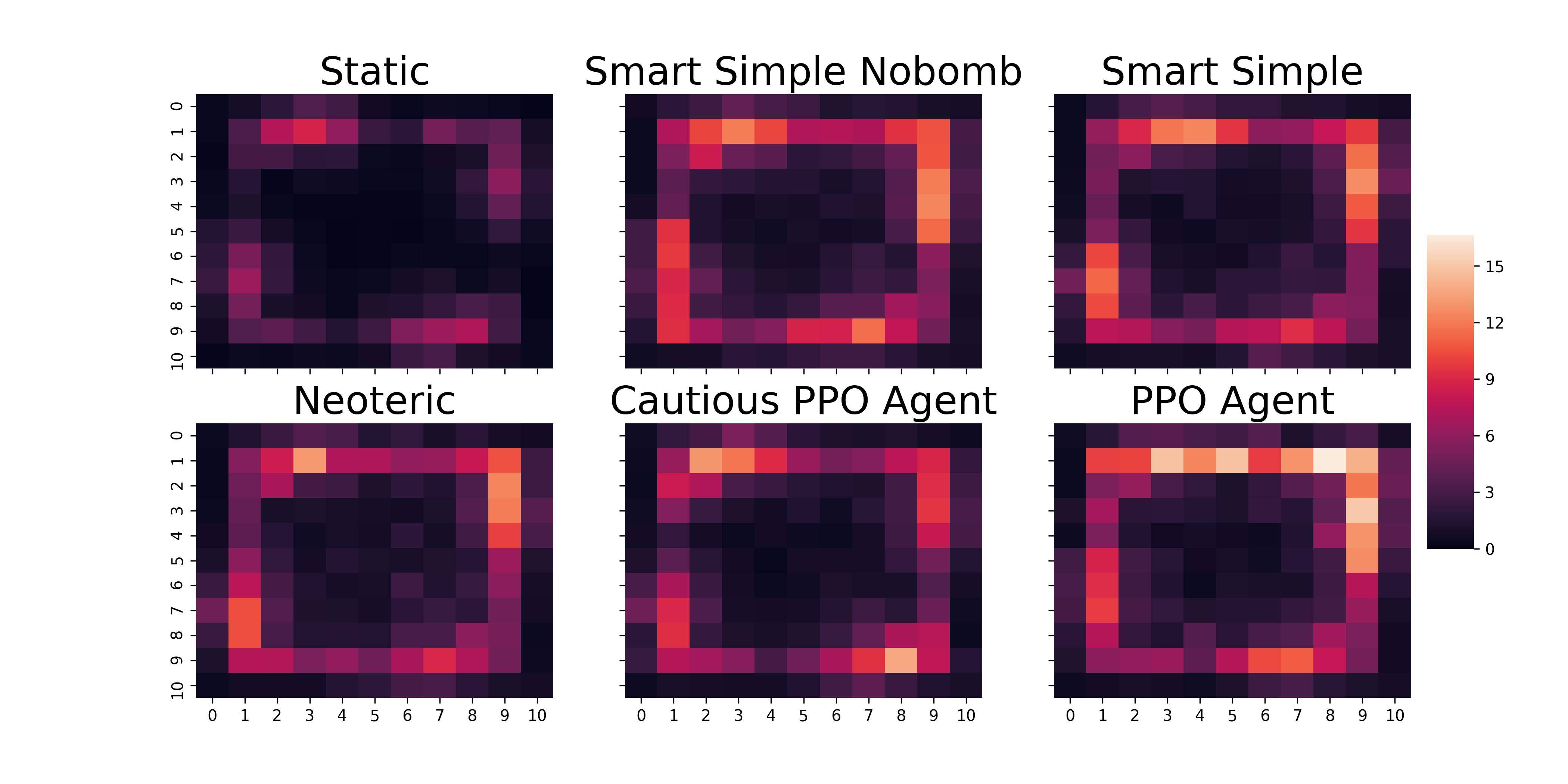}
	\caption{Heatmap of tiles visited by the agent. The values for each tiles are averaged across 1000 games it played against that opponent. The values are summation over the agent and team mate during each game.}
	\label{fig:exploration_heatmap}
\end{figure}

Figure~\ref{fig:exploration_heatmap} shows the heatmap of the 
agents' locations around the board, averaged over 1000 games against each opponent. Brighter areas indicate higher visitation frequency, and the map includes both teammates. Note that the teammates spawn randomly either in the northwest-southeast corners, or in the northeast-southwest corners. Although a static opponent forces the agents to venture out of their quadrants, they can kill it very quickly (as evident from Figure~\ref{fig:game_length_distribution_across_opponents}) and hence heatmap is relatively sparse. PPO-trained opponents force our agents to move all over the board, as they quickly break wooden walls and open up the board, making it hard to create traps. Very high concentration of agents visiting the border is because there is always a passage from one quadrant to another at the borders. Similar analysis for bomb placement is presented in supplementary material.

\begin{figure}[h]
	\begin{subfigure}{.33\textwidth}
	\centering
	\includegraphics[width=\linewidth]{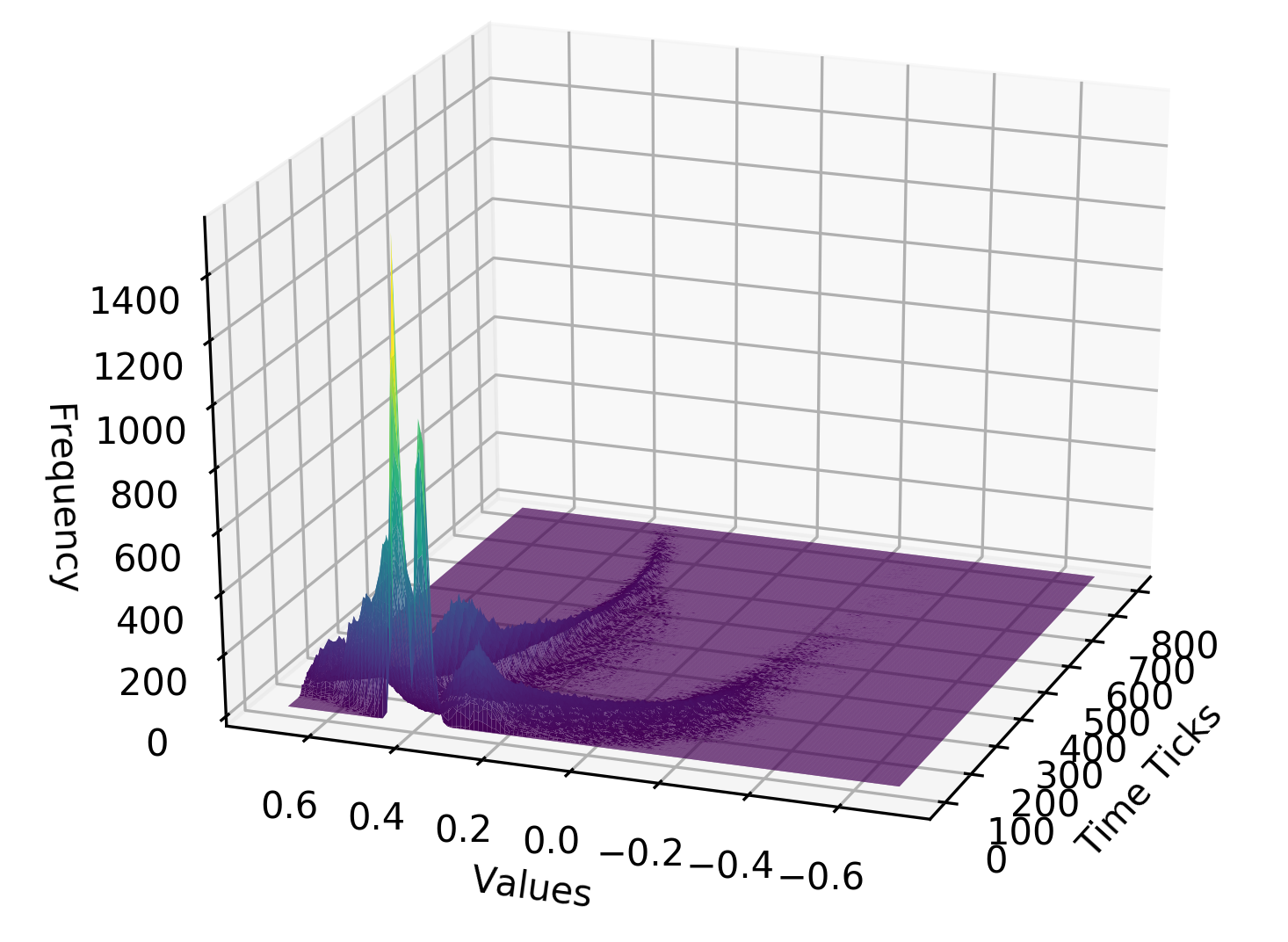}
	\caption{}
	\label{fig:3d_plot_win}
	\end{subfigure}
	\begin{subfigure}{.33\textwidth}
		\centering
		\includegraphics[width=\linewidth]{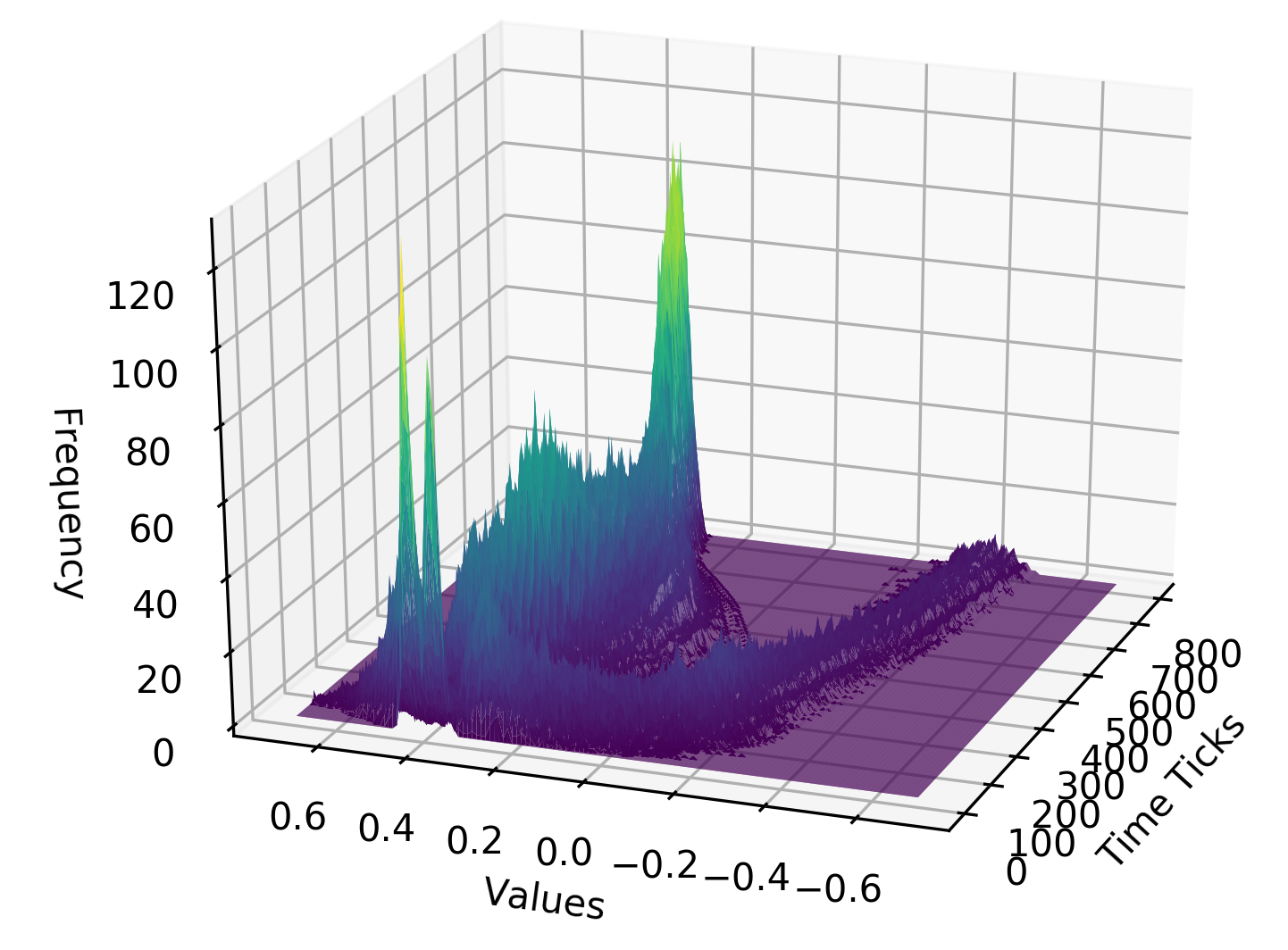}
		\caption{}
		\label{fig:3d_plot_tie}
	\end{subfigure}
	\begin{subfigure}{.33\textwidth}
		\centering
		\includegraphics[width=\linewidth]{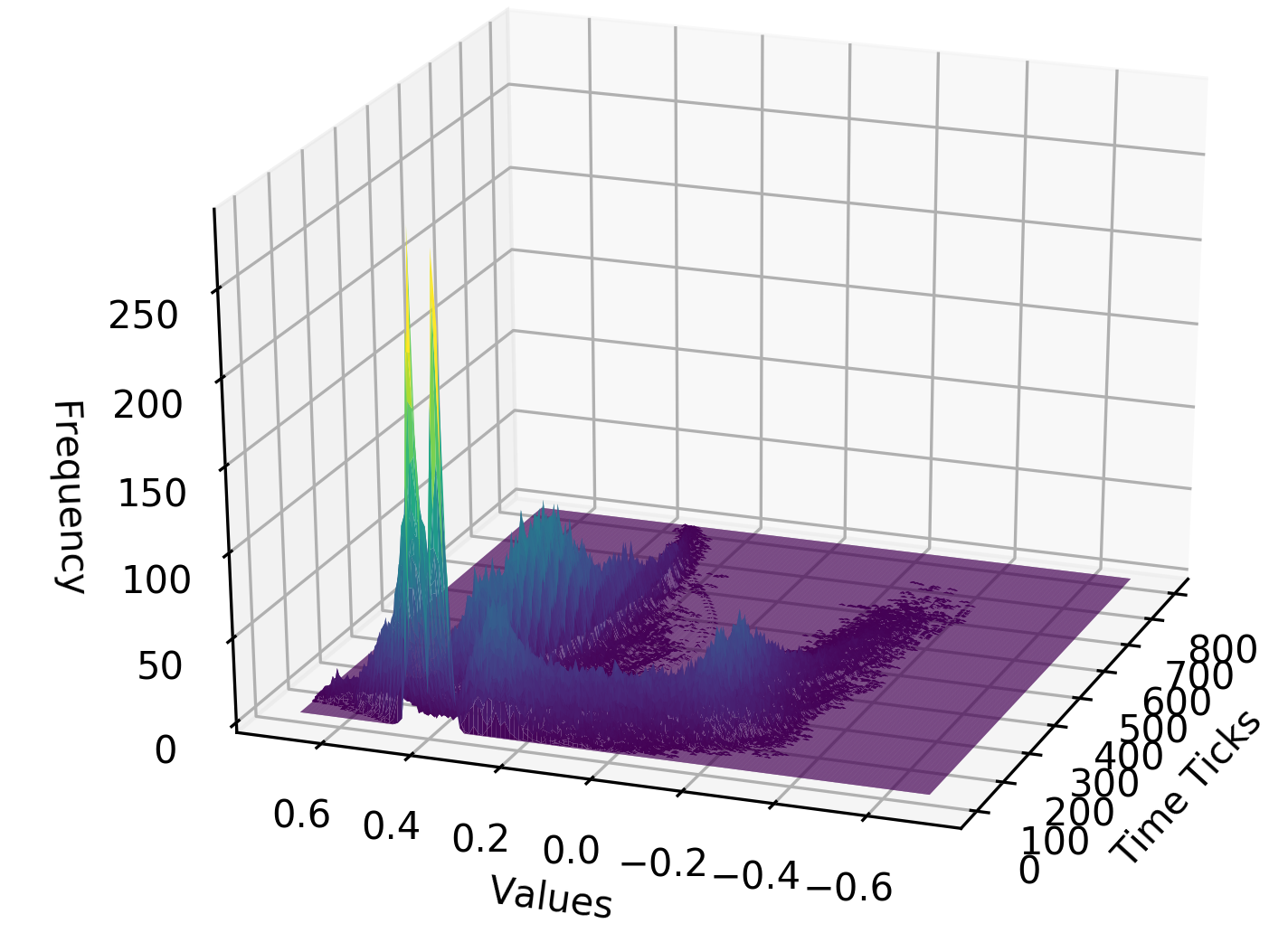}
		\caption{}
		\label{fig:3d_plot_lost}
	\end{subfigure}
	
	\caption{Value prediction graphs, we create the histogram of value prediction values at each time ticks and plot it in a 3D graph, where each slice across the y axis represents a histogram for that time tick. a) when the game was won, b)when the game was tied and c) when the game was lost}
	\label{fig:value_prediction}
	
\end{figure}

Figure~\ref{fig:value_prediction} shows value prediction for state from RL agent for three outcomes of the game. The higher concentration of values on the positive side in the initial phase of the game shows that it is highly optimistic in the start. The peak when the game was won dies down very quickly as the game progress as there are very few games with longer game lengths. When the game was tied, there is are peak at values between 0.4-0.5 reflecting a reward when one enemy agent is killed.  Average value predictions for different opponents also have distinct trajectories. These results are given in supplementary material.

\begin{table}[h!]
	\centering
	\caption{Percentage change after removing the communication and Belief State (SS = Smart Simple, SS n/b = Smart Simple with no bomb placement, and Cau. = Cautious)}
		\begin{tabular}{lrrr||rrr}
			\toprule
			&\multicolumn{3}{c||}{\% change w/o comm} & \multicolumn{3}{c}{\% change w/o belief state}\\
			Opponent    &   Won &   Lost &    Tie &   Won &   Lost &    Tie      \\
			\midrule
			Static              &  96.0 $\rightarrow$ 96.8 &  0.7  $\rightarrow$ 0.3  &  3.3 $\rightarrow$ 2.9 &  $\rightarrow$ 92.4 &  $\rightarrow$ 0.3  &  $\rightarrow$ 7.3\\
			
			SS n/b &  73.2 $\rightarrow$ 70.0 &  16.2 $\rightarrow$ 18.2 & 10.6 $\rightarrow$ 11.8 &  $\rightarrow$ 65.6 &  $\rightarrow$ 18.6 & $\rightarrow$ 15.8\\
			
			SS        &  72.8 $\rightarrow$ 73.3 &  15.4 $\rightarrow$ 16.8 & 11.8 $\rightarrow$ 9.9 & $\rightarrow$ 62.7 &  $\rightarrow$ 21.0 & $\rightarrow$ 16.3\\
			
			Neoteric            &  65.2 $\rightarrow$ 67.8 &  17.2 $\rightarrow$ 17.9 & 17.6 $\rightarrow$ 14.3 &  $\rightarrow$ 59.1 &  $\rightarrow$ 22.0 & $\rightarrow$ 18.9\\
			
			Cau. PPO  &  60.8 $\rightarrow$ 59.4 &  26.3 $\rightarrow$ 27.2 & 12.9 $\rightarrow$ 13.4 &  $\rightarrow$ 50.2 &  $\rightarrow$ 26.4 & $\rightarrow$ 23.4\\
			
			PPO           &  55.8 $\rightarrow$ 55.1 &  22.7 $\rightarrow$ 24.2 & 21.5 $\rightarrow$ 20.1 &  $\rightarrow$ 49.6 &  $\rightarrow$ 24.0 & $\rightarrow$ 26.4\\
			\bottomrule
		\end{tabular}
		\label{tab:results_wo_communication_belief_state}
\end{table}

Table~\ref{tab:results_wo_communication_belief_state} shows change in statistics when 1000 games were played without allowing communication and masking the belief states to zero. We can see that apart from static opponent, losses increase when communication and belief state are unavailable. While removing belief state has significant increase in number of ties as the agent has no global information and generally gets stuck in some state. 
\section{Conclusion}\label{sec:Conclusion}

In a complex game environment such as pommerman, it is really challenging to use any off the shelf RL algorithm. We augment the algorithms with multiple ideas from across the board. We prove that learning in this framework is not only lead to better policy but is also efficient in given compute budget. Our framework wins 65\% of the times against Neoteric agent which was the only agent accessible from TeamRadio competition while winning against Simple Agent reached 80\% which is highest till now from any learning agent. In addition, we also provide multiple perspectives to understanding learnt policy and their possible justification. This environment provides much-needed complexity to develop novel algorithms and frameworks. Learning the communication between the two agents, where they can temporally abstract information flow would be one of the possible future direction.

\bibliography{ref.bib}

\begin{thebibliography}{10}
\providecommand{\url}[1]{#1}
\csname url@samestyle\endcsname
\providecommand{\newblock}{\relax}
\providecommand{\bibinfo}[2]{#2}
\providecommand{\BIBentrySTDinterwordspacing}{\spaceskip=0pt\relax}
\providecommand{\BIBentryALTinterwordstretchfactor}{4}
\providecommand{\BIBentryALTinterwordspacing}{\spaceskip=\fontdimen2\font plus
\BIBentryALTinterwordstretchfactor\fontdimen3\font minus
  \fontdimen4\font\relax}
\providecommand{\BIBforeignlanguage}[2]{{%
\expandafter\ifx\csname l@#1\endcsname\relax
\typeout{** WARNING: IEEEtran.bst: No hyphenation pattern has been}%
\typeout{** loaded for the language `#1'. Using the pattern for}%
\typeout{** the default language instead.}%
\else
\language=\csname l@#1\endcsname
\fi
#2}}
\providecommand{\BIBdecl}{\relax}
\BIBdecl

\bibitem{resnick2018pommerman}
C.~Resnick, W.~Eldridge, D.~Ha, D.~Britz, J.~Foerster, J.~Togelius, K.~Cho, and
  J.~Bruna, ``Pommerman: A multi-agent playground,'' \emph{arXiv preprint
  arXiv:1809.07124}, 2018.

\bibitem{osogami2019real}
T.~Osogami and T.~Takahashi, ``Real-time tree search with pessimistic
  scenarios,'' \emph{arXiv preprint arXiv:1902.10870}, 2019.

\bibitem{zhou2018hybrid}
H.~Zhou, Y.~Gong, L.~Mugrai, A.~Khalifa, A.~Nealen, and J.~Togelius, ``A hybrid
  search agent in pommerman,'' in \emph{Proceedings of the 13th International
  Conference on the Foundations of Digital Games}, 2018, pp. 1--4.

\bibitem{silver2016mastering}
D.~Silver, A.~Huang, C.~J. Maddison, A.~Guez, L.~Sifre, G.~Van Den~Driessche,
  J.~Schrittwieser, I.~Antonoglou, V.~Panneershelvam, M.~Lanctot \emph{et~al.},
  ``Mastering the game of go with deep neural networks and tree search,''
  \emph{nature}, vol. 529, no. 7587, pp. 484--489, 2016.

\bibitem{meisheri2019accelerating}
H.~Meisheri, O.~Shelke, R.~Verma, and H.~Khadilkar, ``Accelerating training in
  pommerman with imitation and reinforcement learning,'' \emph{arXiv preprint
  arXiv:1911.04947}, 2019.

\bibitem{gao2019skynet}
C.~Gao, P.~Hernandez-Leal, B.~Kartal, and M.~E. Taylor, ``Skynet: A top deep rl
  agent in the inaugural pommerman team competition,'' \emph{arXiv preprint
  arXiv:1905.01360}, 2019.

\bibitem{pomm_book_chap}
C.~Resnick, C.~Gao, G.~M{\'a}rton, T.~Osogami, L.~Pang, and T.~Takahashi,
  ``Pommerman {\&} neurips 2018,'' in \emph{The NeurIPS '18 Competition},
  S.~Escalera and R.~Herbrich, Eds.\hskip 1em plus 0.5em minus 0.4em\relax
  Cham: Springer International Publishing, 2020, pp. 11--36.

\bibitem{kartal2018using}
B.~Kartal, P.~Hernandez-Leal, and M.~E. Taylor, ``Using monte carlo tree search
  as a demonstrator within asynchronous deep rl,'' \emph{arXiv preprint
  arXiv:1812.00045}, 2018.

\bibitem{kartal2019safer}
B.~Kartal, P.~Hernandez-Leal, C.~Gao, and M.~E. Taylor, ``Safer deep rl with
  shallow mcts: A case study in pommerman,'' \emph{arXiv preprint
  arXiv:1904.05759}, 2019.

\bibitem{perez2019analysis}
D.~Perez-Liebana, R.~D. Gaina, O.~Drageset, E.~Ilhan, M.~Balla, and S.~M.
  Lucas, ``Analysis of statistical forward planning methods in pommerman,'' in
  \emph{Proceedings of the AAAI Conference on Artificial Intelligence and
  Interactive Digital Entertainment}, vol.~15, no.~1, 2019, pp. 66--72.

\bibitem{malysheva2018deep}
A.~Malysheva, T.~T. Sung, C.-B. Sohn, D.~Kudenko, and A.~Shpilman, ``Deep
  multi-agent reinforcement learning with relevance graphs,'' \emph{arXiv
  preprint arXiv:1811.12557}, 2018.

\bibitem{littman1994markov}
M.~L. Littman, ``Markov games as a framework for multi-agent reinforcement
  learning,'' in \emph{Machine learning proceedings 1994}.\hskip 1em plus 0.5em
  minus 0.4em\relax Elsevier, 1994, pp. 157--163.

\bibitem{bernstein2002complexity}
D.~S. Bernstein, R.~Givan, N.~Immerman, and S.~Zilberstein, ``The complexity of
  decentralized control of markov decision processes,'' \emph{Mathematics of
  operations research}, vol.~27, no.~4, pp. 819--840, 2002.

\bibitem{ng1999policy}
A.~Y. Ng, D.~Harada, and S.~Russell, ``Policy invariance under reward
  transformations: Theory and application to reward shaping,'' in \emph{ICML},
  vol.~99, 1999, pp. 278--287.

\bibitem{openAI_blog}
\BIBentryALTinterwordspacing
Faulty reward functions in the wild. [Online]. Available:
  \url{https://openai.com/blog/faulty-reward-functions/}
\BIBentrySTDinterwordspacing

\bibitem{alphastarblog}
O.~Vinyals, I.~Babuschkin, J.~Chung, M.~Mathieu, M.~Jaderberg, W.~Czarnecki,
  A.~Dudzik, A.~Huang, P.~Georgiev, R.~Powell, T.~Ewalds, D.~Horgan, M.~Kroiss,
  I.~Danihelka, J.~Agapiou, J.~Oh, V.~Dalibard, D.~Choi, L.~Sifre, Y.~Sulsky,
  S.~Vezhnevets, J.~Molloy, T.~Cai, D.~Budden, T.~Paine, C.~Gulcehre, Z.~Wang,
  T.~Pfaff, T.~Pohlen, D.~Yogatama, J.~Cohen, K.~McKinney, O.~Smith, T.~Schaul,
  T.~Lillicrap, C.~Apps, K.~Kavukcuoglu, D.~Hassabis, and D.~Silver,
  ``{AlphaStar: Mastering the Real-Time Strategy Game StarCraft II},'' 2019.

\bibitem{shahmulti}
D.~Shah, N.~Singh, and C.~Talegaonkar, ``Multi-agent strategies for
  pommerman.''

\bibitem{schulman2017proximal}
J.~Schulman, F.~Wolski, P.~Dhariwal, A.~Radford, and O.~Klimov, ``Proximal
  policy optimization algorithms,'' 2017.

\bibitem{peng2018continual}
P.~Peng, L.~Pang, Y.~Yuan, and C.~Gao, ``Continual match based training in
  pommerman: Technical report,'' \emph{arXiv preprint arXiv:1812.07297}, 2018.

\end{thebibliography}
\bibliographystyle{IEEEtran}

\section*{Appendix}

\subsection*{A1 Bomb Placement}

\begin{figure}[h!]
	\centering
	\includegraphics[width=\linewidth]{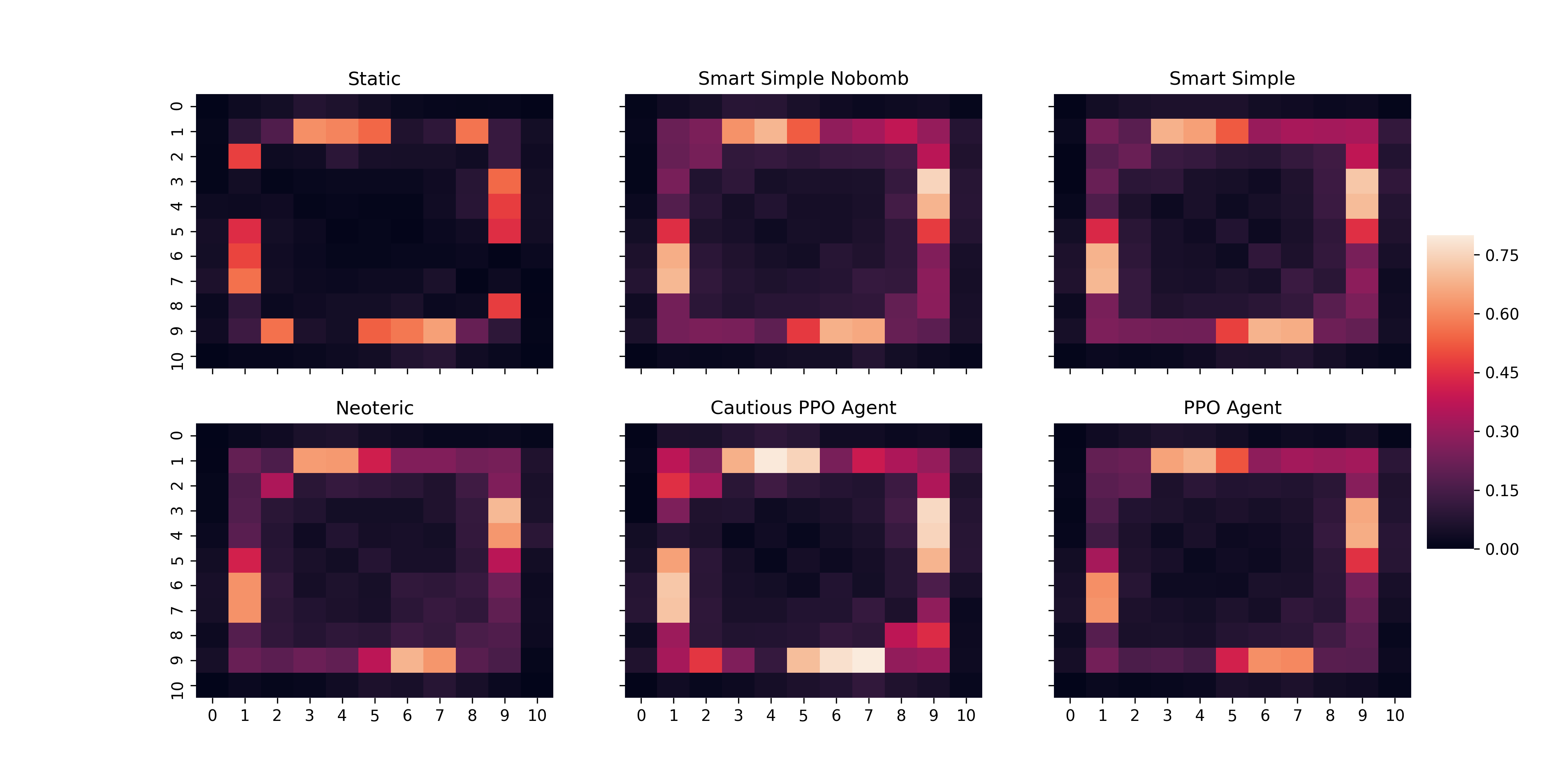}
	\caption{Heatmap of bomb placed at unique tiles on the board. The values for each tiles are averaged across 1000 games it played against that opponent. The values are summation over the agent and team mate during each game.}
	\label{fig:bomb_placement_heatmap}
\end{figure}

To understand the where the agent places a bomb, we plot a heatmap of all the bombs placed by the agents during 1000 games played against each of the opponents as shown in figure~\ref{fig:bomb_placement_heatmap}. 

\subsection*{A2 Value Prediction}

\begin{figure}[h]
	\centering
	\includegraphics[width=0.9\linewidth]{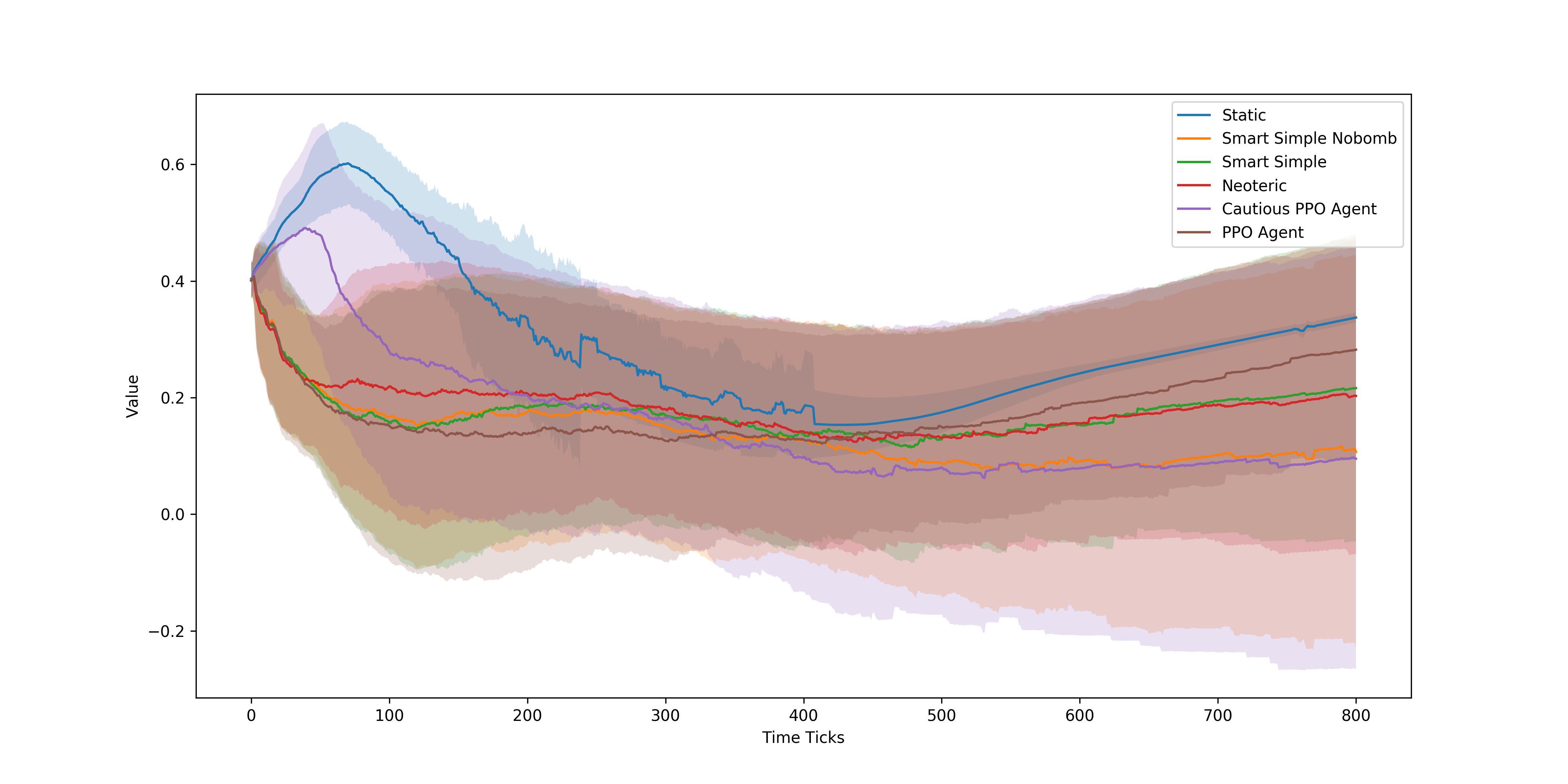}
	\caption{Average value prediction for 1000 games played against each opponent, x-axis is time ticks}
	\label{fig:vpred_across_opp}
\end{figure}

Figure~\ref{fig:vpred_across_opp} shows the average value prediction across different opponent as the game progresses. We can see a clear distinction between the static agent as most of the wins is recorded by $100^{th}$ time tick. In addition, the rest opponents excluding Cautious PPO follow the same trajectory as the game progress. One of the possible reason might it is easier to kill the enemy in the initial part of the game as they can be easy to trap, with wooden walls being present. As the game progresses the board opens up and also enemies acquire kick power and so it becomes harder to kill them. We can observe that the value prediction after $200^{th}$ time ticks almost drops to an average value around 0.2. 

\subsection*{A3 Imitation Learning}

\begin{figure}[h]
	\centering
	\includegraphics[width=\linewidth]{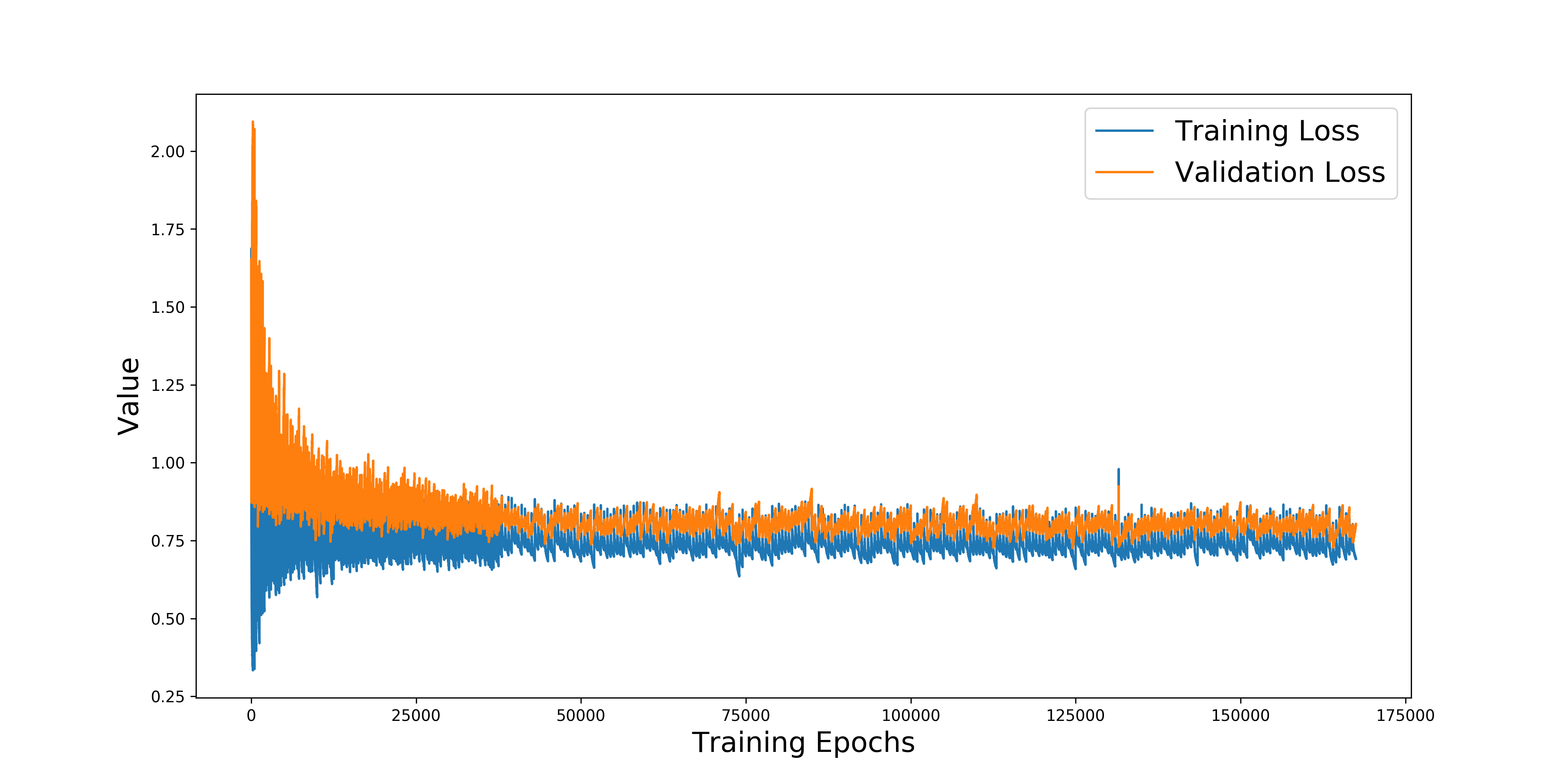}
	\caption{Supervised training and validation set loss while imitation learning, validation split is 20\%}
	\label{fig:imitation_loss}
\end{figure}

We collect 48K samples by playing 12K games with each agent being Smart Simple agent. There is no communication between these agents. Hence we had to augment the samples collected with the communication protocol to generate state representation. We train the network with 4096 batch size, results are shown in figure~\ref{fig:imitation_loss}.

\subsection*{A4 Earlier Training Versions}

\begin{figure}[h]
	\centering
	\includegraphics[width=\linewidth]{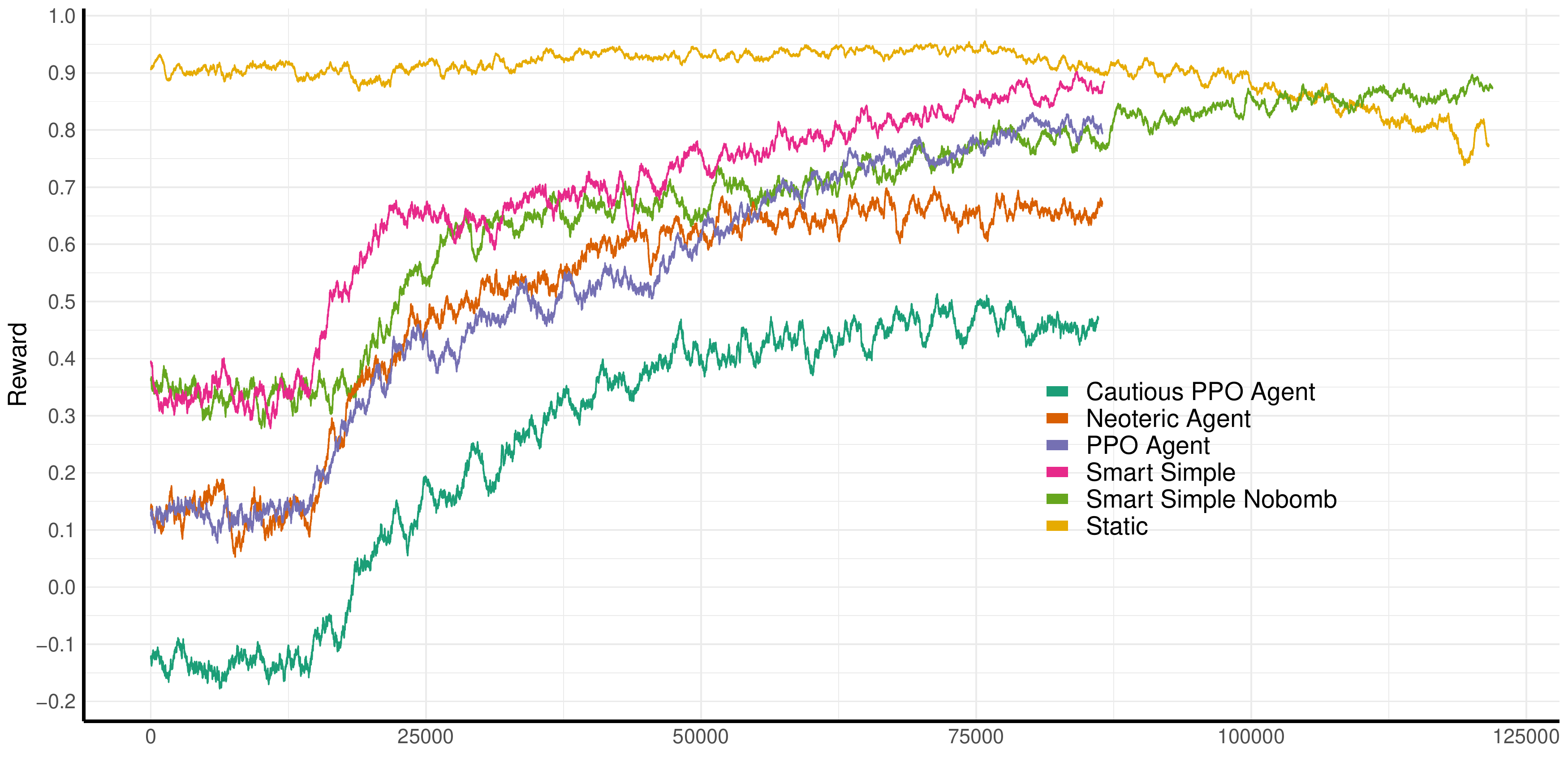}
	\caption{RL training results over Refactored Rewards}
	\label{fig:training_results V2}
\end{figure}

Out initial hypothesis was to equally weight simple agents (Static, Smart Simple Nobomb, smart simple) and more advanced agents (Neoteric, PPO, Cautious PPO) equally. Figure~\ref{fig:training_results V2} shows the training run over $8 \times 10^5$ games. We can see there is a significant increase in rewards against Smart Simple and Smart Simple Nobomb, however, while visually checking it was found that it blockading this heuristic agent diagonally, and the heuristic agent would die by placing a bomb at its own position. RL agent rather than learning robust policy started to exploit this bug. The policy is degrading as can be seen from rewards against other agent decreasing as it is not able to generalize that skill. 

\end{document}